\documentclass[letterpaper, 10 pt, conference]{ieeeconf}  

\IEEEoverridecommandlockouts                              

\usepackage{graphicx} 
\usepackage{wrapfig}
\usepackage{algorithm}
\usepackage[noend]{algpseudocode}
\usepackage[hang,flushmargin]{footmisc}
\usepackage[dvipsnames]{xcolor}
\usepackage{amsmath} 
\usepackage{amssymb}  
\usepackage{cite}
\usepackage{amsmath,bm,subcaption}
\usepackage[font={small},belowskip=-3pt,aboveskip=8pt]{caption}
\usepackage{textcomp}

\usepackage{comment}

\makeatletter
\let\NAT@parse\undefined
\makeatother

\usepackage{hyperref}
\hypersetup{
    colorlinks = true,
    linkcolor={red!50!black},
    citecolor={blue!50!black},
    urlcolor={blue!80!black}
}

\usepackage[noabbrev]{cleveref}
\crefname{equation}{}{}

\newcommand{\etal}{et al.}

\newcommand{\T}{\textsf{T}}
\newcommand{\R}{\mathbb{R}}

\newcommand{\bx}{\bm{x}}
\newcommand{\bz}{\bm{z}}
\newcommand{\bu}{\bm{u}}
\newcommand{\Sb}{\mathcal{S}}
\newcommand{\V}{\mathcal{V}}
\newcommand{\bl}{\bm{l}}
\newcommand{\bK}{\bm{\mathrm{K}}}

\DeclareCaptionFont{blah}{\small\fontseries{n}\fontfamily{mathptmx}\selectfont}
\DeclareMathAlphabet{\mathcal}{OMS}{cmsy}{m}{n}
\DeclareMathOperator*{\argmax}{arg\,max}
\DeclareMathOperator{\atantwo}{atan2}
\DeclareMathOperator*{\argmin}{arg\,min}

\newcommand{\isdefp}{\triangleq}

\renewcommand{\baselinestretch}{0.98}

\title{\LARGE \bf
Simultaneous Control and Trajectory Estimation for Collision Avoidance of Autonomous Robotic Spacecraft Systems*
}

\author{Matthew King-Smith$^{1}$, Panagiotis Tsiotras$^2$,  Frank Dellaert$^{3}$
\thanks{*This work was supported by a contract from the Aerospace Corporation.}
\thanks{$^{1}$M. King-Smith is a Robotics PhD Candidate in the School of Electrical and Computer Engineering, Georgia Institute of Technology, Atlanta, GA, 30332, USA, {\tt\small mcks3@gatech.edu}}%
\thanks{$^{2}$P. Tsiotras is the David \& Andrew Lewis Chair and Professor with the School of Aerospace Engineering, Georgia Institute of Technology, Atlanta, GA, 30332, USA, {\tt\small tsiotras@gatech.edu}}%
\thanks{$^{3}$F. Dellaert is a Professor with the School of Interactive Computing, Georgia Institute of Technology, Atlanta, GA, 30332, USA, {\tt\small frank.dellaert@cc.gatech.edu}}%
}

\begin{document}

\maketitle
\thispagestyle{empty}
\pagestyle{empty}

\begin{abstract}
We propose factor graph optimization for simultaneous planning, control, and trajectory estimation for collision-free navigation of autonomous systems in environments with moving objects. 
The proposed online probabilistic motion planning and trajectory estimation navigation technique generates optimal collision-free state and control trajectories for autonomous vehicles when the obstacle motion model is both unknown and known. 
We evaluate the utility of the algorithm to support future autonomous robotic space missions. 
\end{abstract}

\section{INTRODUCTION}

On-orbit satellite servicing (OSS) holds the promise to refuel,
maintain, upgrade, and repair existing spacecraft, enable
space construction, and actively remove orbital debris~\cite{flores2014review,Gefke2015,Reed2016}. 
During an OSS mission, which involves close proximity operations with other space objects, it is imperative for the servicing spacecraft to be able to adapt to a changing environment while simultaneously achieving the primary mission objective. As such, the vehicle trajectory planning should be implemented online~\cite{JonssonMorrisPedersen2007}, while also considering the combined information (and its uncertainty) collected from onboard system sensors~\cite{Tipaldi2018,vavrina2019safe}. 
In fact, onboard spacecraft autonomy has been identified by the National Aeronautics Space Administration and the European Space Agency as a necessary technology for future space missions~\cite{ Grant,Starek2015}.   

To achieve the desired onboard spacecraft autonomy, there is a need for both online trajectory estimation and planning, given model and sensor uncertainty~\cite{Grant}. 
Traditional online trajectory optimization methods for spacecraft navigation usually solve the optimal control problem~\cite{Halbe2019,Malyuta2020,Capolupo2019}, by performing vehicle trajectory estimation independently from planning~\cite{Tipaldi2018}. 
However, such two-step optimization processes may potentially lead to suboptimal results given that both the estimation and planning problems can be viewed as variants of a single trajectory optimization problem. 

Alternative trajectory optimization approaches, such as sampling-based motion planning (SBMP), offer techniques that combine the planning and estimation into a single optimization problem and have already been applied to spacecraft OSS applications, such as, inspection~\cite{maestrini2020guidance} and exploratory guidance about resident space objects~\cite{Capolupo2017}. 
Additionally, SBMP techniques have been used for planning collision-free spacecraft trajectories in static obstacle environments~\cite{Francis2013}. However, despite SBMP contributions to the spacecraft navigation and estimation communities, all aforementioned SBMP studies do not account for measurement uncertainty from sensors during optimization.

In the robotics community,  probabilistic inference has been used to address various problems related to state estimation~\cite{Dellaert2006,Dellaert2017}, localization~\cite{agha2018slap,barfoot2014batch,Anderson2015,Anderson2015a}, optimal control problems~\cite{kappen2012optimal,ta2014factor,levine2018reinforcement,watson2020stochastic,yang2020equality}, and path planning~\cite{dong2016motion,dong2018sparse,mukadam2017simultaneous} given uncertain sensor measurements and models.
In particular, Mukadam \etal~\cite{Mukadam2018}, have used factor graphs to solve probabilistic inference problems for simultaneous trajectory estimation and planning (STEAP) for robotic systems in static environments. 
STEAP solves the probabilistic estimation and path planning problem in a single factor graph, enabling information to quickly flow between the two problems during online optimization~\cite{Mukadam2018a}. 

Building on this prior work, this research proposes simultaneous control and trajectory estimation (SCATE) for online probabilistic trajectory motion planning and estimation for autonomous satellites in environments with moving obstacles that addresses some of the shortcomings of STEAP, such as, lacking robot dynamics and planning restricted to static obstacle environments. 
Specifically, this work contributes the following advancements over STEAP:
\begin{enumerate}
    \item Adding realistic vehicle dynamics to factor graphs.
    \item Collision-free path planning and trajectory estimation in environments with moving obstacles. 
    \item Evaluation of SCATE navigation on a physical autonomous spacecraft simulator platform~\cite{Cho2009,Tsiotras2014}.
\end{enumerate}

The remainder of the paper is organized as follows: Section~\ref{sect:Approach and Methodology} establishes the mathematics necessary for SCATE trajectory optimization, Section~\ref{sect:Facility} describes the robotic spacecraft platform and testing facility, Section~\ref{sect:Resuts} presents the results of real-time SCATE navigation. Section~\ref{sect:Conclusion} reviews the viability of SCATE navigation for OSS missions, discusses some of its limitations, and suggests some potential avenues for future work.

\section{Approach and Methodology}
\label{sect:Approach and Methodology}

In this section, we review the use of factor graphs for probabilistic inference for trajectory optimization. 
We then introduce STEAP, a factor graph motion planning approach for collision avoidance of kinematic robotic systems in static environments. 
Motivated by STEAP, we propose SCATE, a new algorithm for collision-free navigation of dynamic robotic systems in environments with moving obstacles using factor graphs. 

\subsection{Trajectory Optimization as Probabilistic Inference}  

We represent a trajectory as a continuous-valued function that  maps time $t$ to robot states $\bm{x}(t)$ so as to determine the \textit{maximum a posteriori} (MAP) continuous-time trajectory of $\bm{x}(t)$ given a prior distribution on the space of state trajectories and a likelihood function.  

\subsubsection{Trajectory Prior}

A prior distribution over trajectories can be defined as a vector-valued Gaussian process (GP) $\bm{x}(t) \sim GP(\bm{\mu}(t),\bK(t,t')),$ where $\bm{\mu}(t)$ is the vector-valued mean function and $\bK(t,t')$ is a matrix-valued covariance function. For any collection of times $\bm{t} = \{t_0,\dots,t_N\},\bm{x}$ has a joint Gaussian distribution
$
	\bm{x} \isdefp \left[\bm{x}_0 \cdots \bm{x}_N \right]^\T \sim N(\bm{\mu},\bK),
$
with mean vector $\bm{\mu}$ and covariance kernel $\bK$ defined as
\begin{equation}
	\bm{\mu} \isdefp \left[\bm{\mu}(t_0) \cdots \bm{\mu}(t_N) \right]^\T, 
	\quad \bK \isdefp [\bK(t_i,t_j)] \Big|_{ij,0\le i,j\le N}.
\end{equation}

The prior distribution is then defined by the GP mean $\bm{\mu}$ and the covariance $\bK$ as
\begin{equation}
	p(\bm{x}) \propto \exp \left\{-\frac{1}{2} \|\bm{x} - \bm{\mu} \|^2_{\bK} \right\}. \label{eqn:priorDist}
\end{equation}
Information known \textit{a priori} is then encoded with such priors.

\subsubsection{Likelihood Function}
%
Let $\bz$ be a collection of binary events, where events are defined as actions such as a collision, or receiving a sensor measurement. 
The likelihood function is the conditional distribution $l(\bm{x};\bz) = p(\bz|\bm{x})$, which specifies the probability of events $\bz$ given a trajectory $\bm{x}$. We define the likelihood as a distribution in the exponential family
\begin{equation}
	l(\bm{x};\bz) \propto \exp \left\{-\frac{1}{2} \|\bm{h}(\bx,\bz) \|^2_{\bm{\Sigma}} \right\},\label{eqn:likelihood}
\end{equation}
where $\bm{h}(\bx,\bz)$ is a cost function with covariance matrix $\bm{\Sigma}$.

\subsubsection{Computing the Maximum A Posteriori Trajectory}

Using Bayes rule, we express the posterior distribution of the trajectory given the events in terms of the prior and the likelihood as
$
	p(\bx | \bz) \propto p(\bx)p(\bz|\bm{x}).
$
Then, we can compute the MAP of the set $\Theta~\isdefp~\{ \bx\}$ as
$
	\Theta^* = \argmax_{\Theta}\{p(\bx | \bz)\} = \argmax_{\Theta}\{p(\bx)p( \bz|\bx))\}
	= \argmin_{\Theta}\{- \log(p(\bx)p( \bz|\bx)))\} 
	$
	or
	\begin{equation}   \label{eqn:GeneralMAP}
	\Theta^* 	 = \argmin_{\Theta}  \left\{\frac{1}{2} \|\bm{x} - \bm{\mu} \|^2_{\bK}+\frac{1}{2} \|\bm{h}(\bx,\bz) \|^2_{\bm{\Sigma}} \right\}, 
\end{equation}
where the last expression
follows from~\cref{eqn:priorDist,eqn:likelihood}. 
The underlying sparsity of the problem is computationally exploited by formulating~\cref{eqn:GeneralMAP} into an inference problem on a graphical model~\cite{Dellaert2006}. 

\subsubsection{Factor Graphs for Estimation and Planning}

A computationally efficient way to compute the MAP trajectory given in~\cref{eqn:GeneralMAP} is to exploit the known structure of the problem by representing the posterior distribution as a \textit{factor graph}. 
As shown in~\cite{Kschischang2001}, a factor graph allows for any distribution to be \textit{factored} into a product of functions that is organized as a bipartite graph $G =\{ \Theta,\Phi,E \}$. 
The graph consists of factor nodes $\Phi\isdefp\{\phi_0, \dots, \phi_V\}$,  variable nodes given by the set $\Theta~\isdefp~\{ \bx \}$, and edges $E$ which connect the two types of nodes, as shown in~\Cref{fig:STEAP}.

Letting $\Theta_i$ be a variable subset of $\Theta$, then the posterior distribution can be expressed as the product of the factors
\begin{equation}
	p(\bx) \propto \prod_{i=0}^{V}\phi_i(\Theta_i).
\end{equation} 
Factor graphs that are sparse lend to sparse precision matrices, which are exploitable, yielding a computationally efficient way to determine $\Theta^*$ from~\cref{eqn:GeneralMAP}~\cite{Dellaert2006}.

\subsection{Simultaneous Trajectory Estimation and Planning}

\begin{figure}[h]
	\vspace*{-5pt}
	\centering
	\includegraphics[width=1\linewidth]{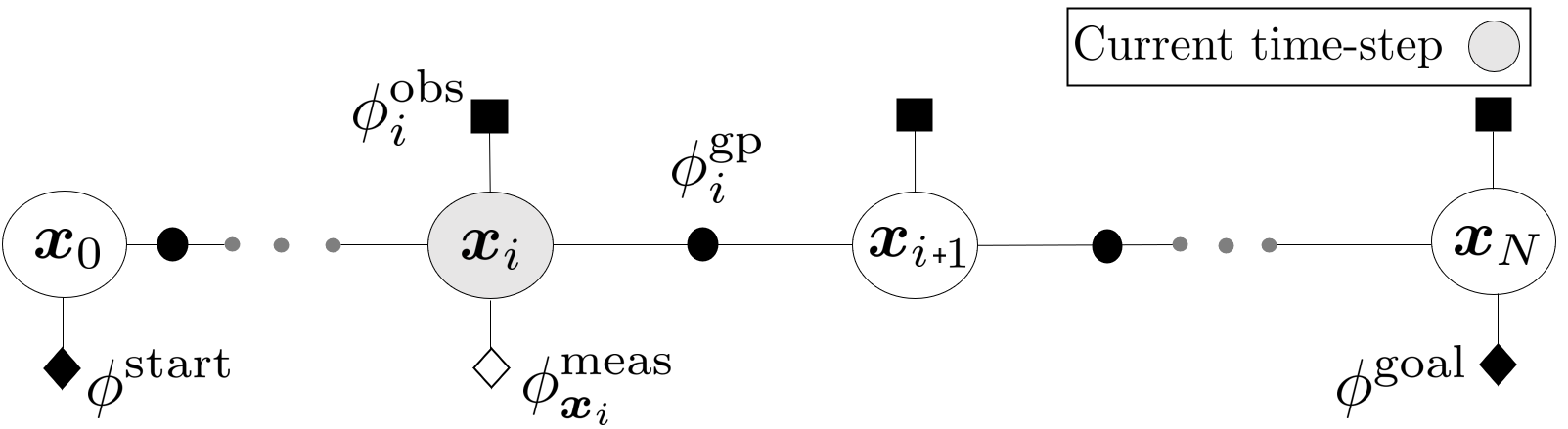}
	\caption{STEAP factor graph for trajectory estimation and planning.}
	\label{fig:STEAP}
\end{figure}

The simultaneous trajectory estimation and planning (STEAP) algorithm proposed by Mukadam \etal~\cite{Mukadam2018}, is a unified probabilistic framework constructed via factor graphs for both past state estimation and future path planning of robotic systems. The state trajectory, $\bx$, is represented by the GP prior given in~\cref{eqn:priorDist}, with mean and covariance functions $\bm{\mu},\bK$, given all sensor data and cost information collected into a single likelihood. 
The posterior distribution is thought to represent events that happen in the past and in the future simultaneously and is represented by
\begin{equation}
	p(\hat{\bx} \cup \check{\bx} | \bz) = \phi^{\rm gp}\phi^{\rm meas}\phi^{\rm obs}\phi^{\rm fix},
	\label{eqn:STEAP}
\end{equation}
where $\phi^{\rm gp},\phi^{\rm meas},\phi^{\rm obs},\phi^{\rm fix},$ are the GP prior, state measurement, obstacle, and goal factors, respectively, of the graph, as shown in~\Cref{fig:STEAP}, and are defined in Section~\ref{section:FactorDefs}.

The MAP solution to the factor graph formulation given in~\cref{eqn:STEAP}, i.e.,  $\Theta^* = \{\hat{\bx} \cup \check{\bx}\}$, solves both the estimation (i.e., $\hat{\bx}$) and planning (i.e., $\check{\bx}$) problems in a single step. As the robot transverses the trajectory over time, new measurements and cost information is used to appropriately update the likelihood and graph for online planning and estimation~\cite{Mukadam2018a}. 

STEAP, however, only considers the kinematics for robot systems that operate in static object environments.
Furthermore, STEAP generates GP priors by linear time-varying stochastic differential equations maintaining constant velocity, i.e., noise-on-acceleration input~\cite{dong2018sparse,Mukadam2018}. 
Such GP priors factor according to 
$	\phi^{\rm gp} = \prod_i \phi_i^{\rm gp}(\bx_i,\bx_{i+1}),$
where any GP prior factor connects to only its two neighboring states (without control), forming a (Gauss-Markov) chain.

\subsection{Simultaneous Control and Trajectory Estimation}

To account for environments with moving obstacles, we must also consider the robot (spacecraft) dynamics. 
Hence, we propose simultaneous control and trajectory estimation (SCATE) to determine the MAP solution to the set $\Theta~\isdefp~\{ \bx,\bu,\bl \}$, where $\bu$ is the control input trajectory, and $\bl$ is the set of obstacle-$\bl$ locations. 
We investigate both \textit{reactive} and \textit{predictive} SCATE navigation given an unknown and known obstacle motion model, respectively. 
Finally, the MAP solution to SCATE factor graphs is given as $\Theta^*~\isdefp~\{ \hat{\bx} \cup \check{\bx} ,\hat{\bu} \cup \check{\bu},\hat{\bl} \}$, where $\hat{\bu},\hat{\bx},\hat{\bl}$ are estimates of $\bu,\bx,\bl$, and $\check{\bx},\check{\bu}$ are the planned state and control input, respectively, to execute along the remaining trajectories.  

\begin{figure}[h]
	\vspace*{-6pt}
	\centering
	\includegraphics[width=.99\linewidth]{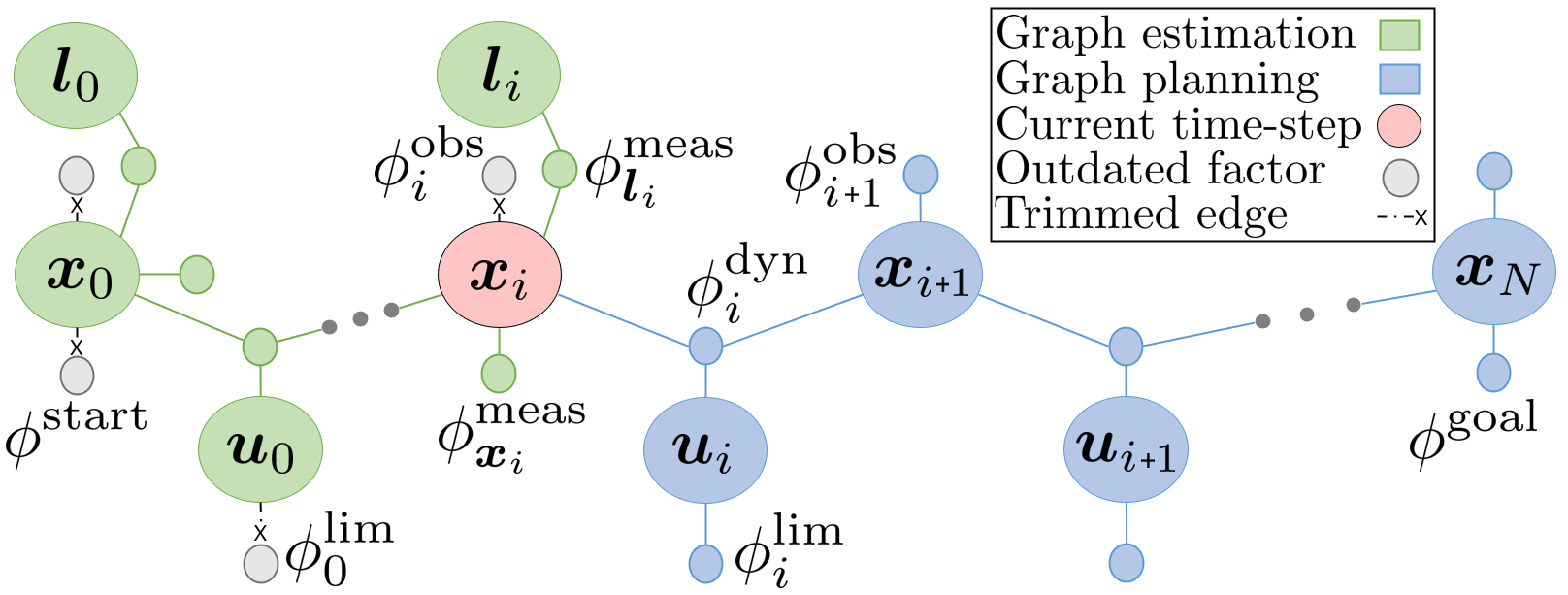}
	\caption{SCATE factor graph for control and trajectory estimation.}
	\label{fig:reactiveSCATE}
\end{figure}

Reactive SCATE factor graphs are given by the posterior distribution 
\begin{equation}
	p(\hat{\bx} \cup\check{\bx},\hat{\bu} \cup \check{\bu}, \hat{\bl} | \bz ) = \phi^{\rm dyn}\phi^{\rm meas}\phi^{\rm lim}\phi^{\rm obs}\phi^{\rm fix},
	\label{eqn:reactiveSCATE}
\end{equation} 
where $\phi^{\rm dyn},\phi^{\rm lim}$ are dynamic, and control limit factors, respectively, as shown in~\Cref{fig:reactiveSCATE}, and are defined in Section~\ref{section:FactorDefs}.
Reactive SCATE factor graphs plan for state $\bx_i$ assuming obstacles observed are static in the environment. 

Predictive SCATE factor graphs, are given the obstacle trajectory, $\tilde{\bl}$, and are expressed by the posterior distribution
\begin{equation}
	p(\hat{\bx} \cup\check{\bx},\hat{\bu} \cup \check{\bu}, \hat{\bl} | \bz,\tilde{\bl}) = \phi^{\rm dyn}\phi^{\rm meas}\phi^{\rm lim}\phi^{\rm obs}\phi^{\rm fix}.
	\label{eqn:PredictiveSCATE}
\end{equation} 
Knowing $\tilde{\bl}$ in advance enables predictive SCATE to allocate cost to future obstacle locations via the remaining obstacle factors, i.e., the set $\{\phi_{i+1}^{\rm obs},\dots,\phi_{N}^{\rm obs} \}$, during planning for $\bx_i$.
\begin{figure*}[!h]
	\begin{subfigure}[h]{0.49\linewidth}
		\centering
		\vspace*{8pt}
		\includegraphics[width=0.95\linewidth]{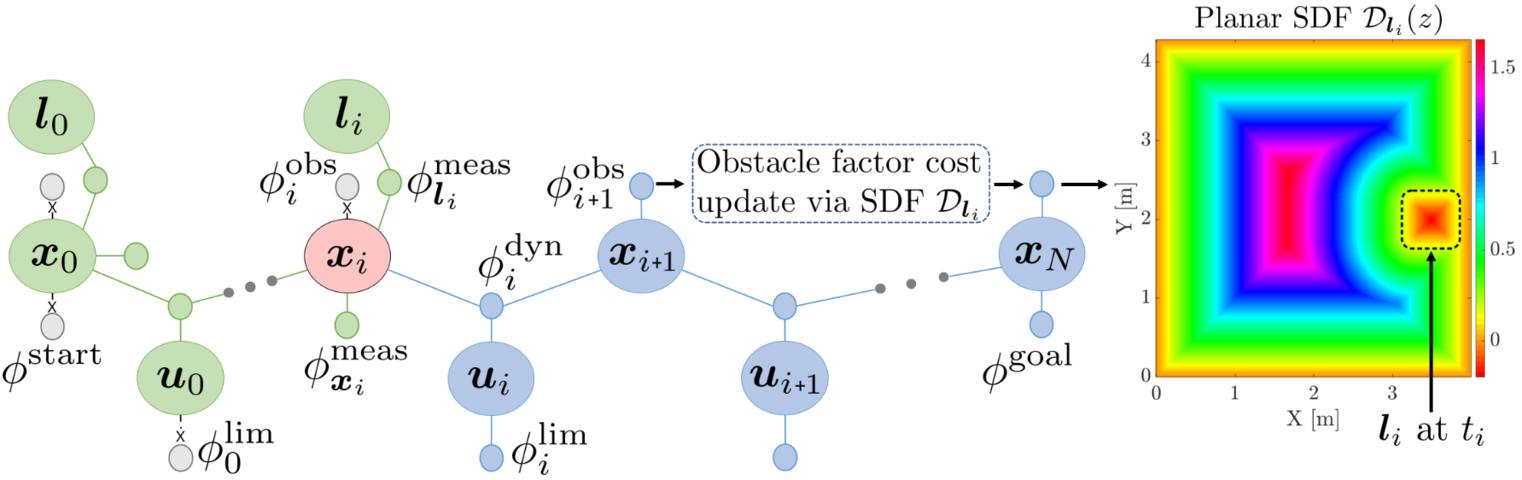}
		\vspace*{4pt}
		\caption{Reactive: continuous factor cost (re)assignment using $\mathcal{D}_{\bl_i}(z)$.}
		\label{fig:reactiveObsAss}
	\end{subfigure}%
	\hfill
	\begin{subfigure}[h]{0.49\linewidth}
	    \vspace*{3pt}
		\includegraphics[width=0.95\linewidth]{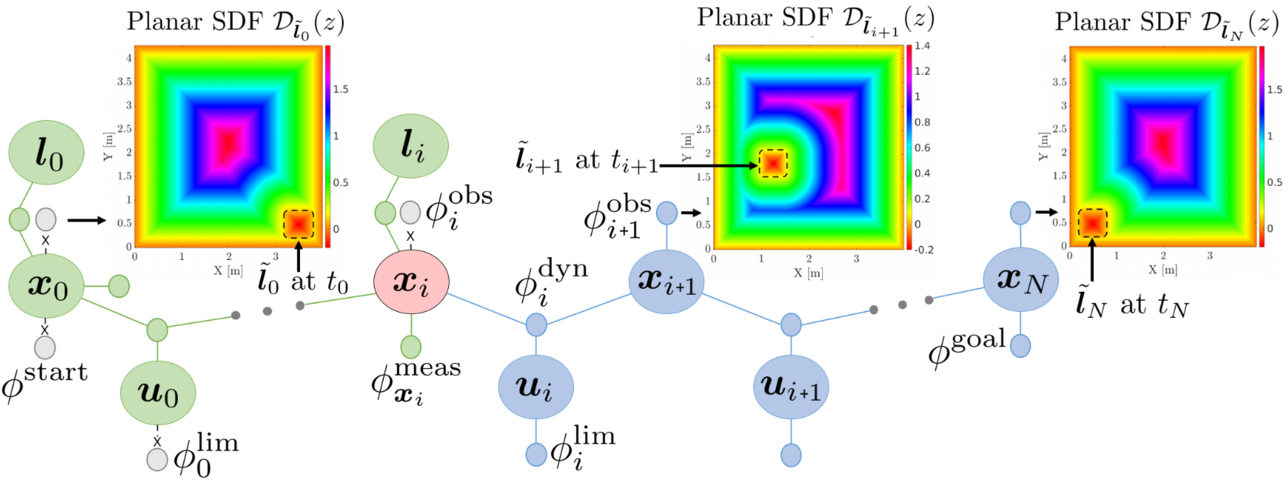}
		\caption{Predictive: preemptive factor cost assignment using $\{\mathcal{D}_{\tilde{\bl}_0},\dots,\mathcal{D}_{\tilde{\bl}_N}\}$.}
		\label{fig:PredObsAss}
	\end{subfigure}
	\caption{SCATE obstacle factor cost assignment. The color bar indicates the value of SDF $\mathcal{D}_{\bl_i}$ at a point $z$.}
	\label{fig:SCATEObsAss}
\end{figure*}

\Cref{fig:reactiveSCATE} also shows the removal of outdated planning factors, i.e., the set $\{\phi^{\rm start},\phi_{0}^{\rm lim},\ldots,\phi_{i}^{\rm obs}\}$, from the graph such that they do not influence the $t_i$-th optimization.

\subsubsection{SCATE Factor Definitions}
\label{section:FactorDefs}

We now define the factors $\phi^{\rm fix},\phi^{\rm meas},\phi^{\rm obs},\phi^{\rm dyn},\phi^{\rm lim}$ given in~\cref{eqn:STEAP,eqn:reactiveSCATE,eqn:PredictiveSCATE}.

\noindent \textbf{Start and goal factor:} 
The multivariate Guassian factors
\begin{align}
	\phi^{\rm start}(\bx_0) &\propto \exp \left\{ -\frac{1}{2}\|\bx_0 - \bx_{\rm start} \|^2_{\bm{\Sigma}_{\rm fix}}   \right\}, \label{eqn:StartFactor}\\
	\phi^{\rm goal}(\bx_N) &\propto \exp \left\{ -\frac{1}{2}\|\bx_N - \bx_{\rm goal} \|^2_{\bm{\Sigma}_{\rm fix}}   \right\},
	\label{eqn:EndFactor}
\end{align}
with the mean as the start or goal and a small covariance $\bm{\Sigma}_{\rm fix}$ define $\phi^{\rm fix}=\phi^{\rm start}(\bx_0)\phi^{\rm goal}(\bx_N)$ and root the trajectory endpoints to the start and goal locations. 

\noindent \textbf{Measurement factors:} 
For simplicity, we use multivariate Gaussian measurement factors for state measurements
\begin{equation}
	\phi_{\bx_i}^{\rm meas}(\bx_i) \propto \exp \left\{- \frac{1}{2} \| \bx_i - \bm{z}_{\bx_i}^{\rm meas}\|_{\bm{\Sigma}^{\bx}_{\mathrm{meas}}}^2   \right\},
	\label{eqn:MeasFactors}
\end{equation}
where, with a slight abuse of notation, $\bm{z}_{\bx_i}^{\rm meas}$ denotes the $i$-th state measurement with covariance $\bm{\Sigma}^{\bx}_{\mathrm{meas}}$. 
Assuming an obstacle is observed via relative bearing and range measurements,  $\bz^{\rm meas}_{\theta_i},\bz^{\rm meas}_{r_i}$, with covariance matrices $\bm{\Sigma}^{\theta}_{\mathrm{meas}},\bm{\Sigma}^{r}_{\mathrm{meas}}$, respectively,  we can then define a bearing and range factors as
\begin{equation}
\begin{aligned}
	\phi_{\theta_i}^{\rm meas}(\bx_i,\bl_i) \propto
	\exp& \left\{    - \frac{1}{2}   \| h_{\theta_i}(\bx_i,\bl_i) - \bm{z}_{\theta_i}^{\rm meas} \|_{\bm{\Sigma}^{\theta}_{\mathrm{meas}}}^2 \right\}, \\
	\phi_{r_i}^{\rm meas}(\bx_i,\bl_i) \propto
	\exp& \left\{    - \frac{1}{2} \| h_{r_i}(\bx_i,\bl_i) - \bm{z}_{r_i}^{\rm meas}\|_{\bm{\Sigma}^{r}_{\mathrm{meas}}}^2 
	 \right\},
	\label{eqn:MeasObstacleFactor}
\end{aligned}
\end{equation}
where $h_{\theta_i}(\bx_i,\bl_i) = \atantwo (\bl_{i,y}-\bx_{i,y},\bl_{i,x}-\bx_{i,x}),$ and $h_{r_i}(\bx_i,\bl_i)~=~\|\bm{r}_{\mathrm{ B/ A}}^\mathcal{S}\|_2$ is the relative range given the planar displacement vector between the robot and obstacle, $\bm{r}_{\mathrm{ B/ A}}^\mathcal{S}~=~{\small \begin{bmatrix}
	\mathrm{c}(\bx_{i,\psi^{\mathcal V/ S}}) & -\mathrm{s}(\bx_{i,\psi^{\mathcal V/ S}}) \\
	\mathrm{s}(\bx_{i,\psi^{\mathcal V/ S}})  & \mathrm{c}(\bx_{i,\psi^{\mathcal V/ S}}) \\
\end{bmatrix}^\T \begin{bmatrix}\bl_{i,x}-\bx_{i,x}\\ \bl_{i,y}-\bx_{i,y}\end{bmatrix}}$, where $\mathrm{c}(\cdot)=\cos(\cdot)$, and $\mathrm{s}(\cdot)=\sin(\cdot)$, as shown in~\Cref{fig:Kinematics}. A 2D bearing range factor is then given by $\phi^{\mathrm{meas}}_{\bl_i} = \phi_{\theta_i}^{\rm meas}\phi_{r_i}^{\rm meas}$.


\noindent \textbf{Linear time-invariant dynamics factor:} We assume that the dynamics of the robotic system are governed by a linear time-invariant (LTI) state-space model, i.e.,
\begin{equation}   	\label{eqn:SimpleSS}
	\dot{\bx}(t) = A \bx(t) + B \bu(t),
\end{equation}
which, after discretization with
$\delta t = t_{i+1} - t_{i}$, yields
\begin{equation}
	\bx_{i+1} = F_x \bx_i + F_u\bu_i,
	\label{eqn:SS_prop}
\end{equation}
where $F_x = e^{A\delta t}$ and $F_u = e^{A\delta t}\int_{t_{i}}^{t_{i+1}} e^{-A \tau}\mathrm{d}\tau B$.
Given~\cref{eqn:SS_prop}, we define the discrete LTI dynamics factor as 
\begin{equation}
	\phi_i^{\rm dyn}(\bx_{i+1},\bx_{i},\bu_i) \hspace{-2pt} \propto \hspace{-2pt} \exp\left\{ \hspace{-2pt} -\frac{1}{2} \|\bx_{i+1} - F_x \bx_i - F_u \bu_i \|^2_{\bm{\Sigma}_{\rm dyn}} \hspace{-2pt} \right\}\hspace{-3.5pt},
	\label{eqn:LTIfactor}
\end{equation}
where $\bm{\Sigma}_{\rm dyn}$ is the covariance matrix of the factor. 

\noindent \textbf{Control limit factor:} 
The control limit factor is designed to ensure that control trajectories respect limits. Defining the hinge loss cost function for the $\bu_i $ control input
\begin{align}
&\bm{h}(\bu_i) = \nonumber \\
&\left[
\begin{cases}
	u^j_{i,-} + u^j_{i,ths} - u^j_i & \mathrm{if} \ u^j_i < u^j_{i,-} + u^j_{i,ths} \\
	0 & \mathrm{if} \ u^j_i \le u^j_{i,+} - u^j_{i,ths} \\
	u^j_i -u^j_{i,+} + u^j_{i,ths}  & \mathrm{otherwise} 
\end{cases} \right] \Bigg|_{1 \le j \le m},
\label{eqn:HingeLossLim}
\end{align}
where $u^j_{i,-},u^j_{i,+}$ are the lower and upper limit of the $u_i^j$ component of $\bu_i$, respectively, and $u^j_{i,ths}$ is a threshold value, then given~\cref{eqn:HingeLossLim}, the control limit factor is given as
\begin{equation}
	\phi_i^{\rm lim}(\bu_{i}) \propto \exp\left\{ -\frac{1}{2} \|\bm{h}(\bm{u}_i) \|^2_{\bm{\Sigma}_{\rm lim}}\right\},
	\label{eqn:ControlLimitFactor}
\end{equation}
where $\bm{\Sigma}_{\rm lim}$ is the covariance matrix of the factor.

\noindent \textbf{Obstacle factor:} 
\label{sect:ObstacleFactor} 
The obstacle factor derives from the likelihood function, which is given by the conditional distribution $l_{\rm obs}(\bm{x}_i;\bz_i=0) = p(\bz_i=0|\bm{x}_i)$, which specifies the probability of being clear of collisions, or the probability that collision event $\bz_i=0$, given the current configuration $\bx_i $. 
This likelihood is represented as a distribution in the exponential family
\begin{equation}
	\phi_{i}^{\rm obs}(\bm{x}_i) = l_{\rm obs}(\bm{x}_i;\bz_i=0) \propto \exp \left\{-\frac{1}{2}\|\bm{h}_{\bl_i}(\bx_i) \|^2_{\bm{\Sigma}_{\rm obs}}\right\}, \label{eqn:ObsLike}
\end{equation}
where $\bm{h}_{\bl_{i}}(\bx_i)$ is a vector-valued \textit{obstacle cost function} with embedded obstacle location $\bl_i \in \R^q$ and $\bm{\Sigma}^{-1}_{\rm{obs}} = \sigma_{\rm obs}\bm{I}$ is a hyperparameter~\cite{barfoot2014batch}. 

Given the likelihood in~\cref{eqn:ObsLike}, we define the hinge loss~\footnote{The hinge loss is not differentiable at $d(z)=\epsilon$, so in this implementation $c'(z)=-0.5$ when $d(z)=\epsilon$~\cite{dong2016motion}.}
\begin{equation}
	\bm{c}(z,\bl_i) = \begin{cases}
		-\mathcal{D}_{\bl_i}(z)+\epsilon, & \mathrm{if} \ \mathcal{D}_{\bl_i}(z)\le \epsilon ,\\
		0, & \mathrm{if} \ \mathcal{D}_{\bm{l}_{i}}(z) > \epsilon,\\
	\end{cases}
\end{equation}
where $\mathcal{D}_{\bl_i}(z)$ is the \textit{signed distance field} (SDF) about the point $z$
given obstacle-$\bl_i$, and provides the distance from point $z$ in the workspace to the closest obstacle surface, and $\epsilon$ is a ``safety distance'' indicating the boundary of the ``danger area'' near obstacle surfaces, as shown in~\Cref{fig:SCATEObsAss}.

For fast collision checking, we adopt sphere representation of $M$-body systems~\cite{Zucker2013} . 
The obstacle cost function for each state $\bx_i$ is determined by computing the signed distance for a sphere representing each body, i.e., $s_j (j=1,\dots,M$ bodies), and collecting them into a single vector such that 
\begin{equation}
	\bm{h}_{\bl_i}(\bx_i) = \left[\bm{c}(\bm{f}(\bx_i,s_j),\bl_i)\right] \Big|_{{1\le j \le M}},
\end{equation}
where $\bm{f}(\cdot)$ maps the state $\bx_i$ to the corresponding set of sphere locations in the workspace (more details in~\cite{dong2016motion}).

\Cref{fig:SCATEObsAss} shows how assignment of obstacle factors differs between reactive and predictive SCATE factor graphs. In~\Cref{fig:reactiveObsAss}, reactive SCATE factor graph assigns future obstacle factors, i.e., the set $\{\phi_{i+1}^{\rm obs},\dots,\phi_{N}^{\rm obs} \}$, assuming obstacle, $\bl_i$, is static in the environment with SDF $\mathcal{D}_{\bl_i}(z)$.
In~\Cref{fig:PredObsAss}, predictive SCATE factor graphs use $\tilde{\bl}$ embedded into a sequence of SDFs, i.e., the set $\{\mathcal{D}_{\tilde{\bl}_0},\mathcal{D}_{\tilde{\bl}_1},\dots,\mathcal{D}_{\tilde{\bl}_N}\}$, such that the obstacle factor assignment reflects the anticipated obstacle trajectory. 
Finally, \Cref{SCATE_alg} shows that predictive SCATE factor graphs do not need to reassign cost for future obstacles and hence require fewer computational steps than reactive SCATE factor graphs for the same problem. 

\subsubsection{Computational Complexity Analysis}

We solve for $\Theta_i^*$ at time-step $t_i$ by linearizing~\cref{eqn:GeneralMAP}, given~\cref{eqn:reactiveSCATE} or~\cref{eqn:PredictiveSCATE}, about a trajectory (starting with guess $\tilde{\bx},\tilde{\bu}$ at $t_0$) and then use variable elimination~\cite{blair1993introduction} to solve ($n + m)N + iq$ local linear sub-problems~\cite{Dellaert2006}. Since finding an optimal elimination order is NP-complete~\cite{Yannakakis1981}, we follow~\cite{Pradhan2021OptimalCF}, and use Column Approximate Minimum Degree (ColAMD)~\cite{Davis2004}. When eliminating variables in the estimation portion of SCATE factor graphs, because the number of constraints per pose is constant, the complexity is $O(1)$~\cite{Kaess2011}. Elimination of planning state and control variables results in $O(n^3),O(m^3)$ complexity per sub-problem, respectively~\cite{yang2020equality}, but since the ColAMD algorithm has complexity $O(E)$ for a bounded degree graph with $E$ edges~\cite{Amestoy1996}, then the total complexity is $O(i + (N-i)\cdot(\kappa_1 n^3 + \kappa_2 m^3) )$ for $\kappa_1,\kappa_2 >0$. 

\subsubsection{Pseudocode for Computer Implementation of SCATE}

\noindent\begin{minipage}{\linewidth}
\vspace*{-8pt}
\renewcommand\footnoterule{} 
\begin{algorithm}[H]
\caption{SCATE Factor Graph Optimization}
\label{SCATE_alg}
\footnotesize
\begin{algorithmic}[1]
\Statex \textbf{Generate Initial Plan:}	
\item $G$ = $\mathsf{gtsam}$.NonlinearFactorGraph() \Comment{Create graph via $\mathsf{gtsam}$~\cite{dellaert2012factor}}
\State $G.\Phi.\mathsf{add}\left(\phi^{\rm fix}\right)$ \Comment{Root graph endpoints with factors in~\cref{eqn:StartFactor,eqn:EndFactor}}
\For{$i = 0:N$} \Comment{Add constraint factors to graph over $\{t_0,\dots,t_N\}$}
	\If{$i <N$}
		\State $G.\Phi.\mathsf{add}\left(\phi_i^{\rm dyn}\left(\bx_{i+1},\bx_{i},\bu_{i}\right)\right)$ \Comment{Dynamics factor in~\cref{eqn:LTIfactor}}
		\State $G.\Phi.\mathsf{add}\left(\phi_i^{\rm lim}\left(\bu_i\right)\right)$ \Comment{Control limit factor in~\cref{eqn:ControlLimitFactor}}
	\EndIf
	\If{$PlanningMode$ = Reactive}  \Comment{Plan without $\tilde{\bl}$}
	\State $G.\Phi.\mathsf{add}\left(\phi_i^{\rm obs}\left(\bx_i\right),\emptyset \right)$\footnote{Obstacle factors encode an empty bounded workspace when $\bl_i = \emptyset$.} \Comment{Obstacle factor in~\cref{eqn:ObsLike}}
	\ElsIf{$PlanningMode$ = Predictive} \Comment{Plan given $\tilde{\bl}$}
	\State $G.\Phi.\mathsf{add}\left(\phi_i^{\rm obs}\left(\bx_i\right),\tilde{\bl}_i\right)$ \Comment{Obstacle factor in~\cref{eqn:ObsLike}}
	\EndIf
\EndFor
\item $\Theta_{0} = \mathsf{gtsam}$.Values($\tilde{\bx},\tilde{\bu}$) \Comment{Initialize solution guess given $\tilde{\bx},\tilde{\bu}$ at $t_0$}
\item $\Theta^*_{0} \gets  \mathsf{gtsam}$.LevenbergMarquardt($G,\Theta_{0}$).Optimize() \Comment{Find solution}
\Statex
\Statex \textbf{Iterative Online Factor Graph Optimization:}
\For{$i = 0:N$} \Comment{Iteratively solve factor graph along $\{t_0,\dots,t_N\}$}
	\State $G.\Phi.\mathsf{add}\left(\phi_{\bx_i}^{\rm meas}(\bx_i),\phi_{\bl_i}^{\rm meas}(\bx_i,\bl_i)\right)$ \Comment{Factors in~\cref{eqn:MeasFactors,eqn:MeasObstacleFactor}}
	\State $\Theta^*_i.\mathsf{add}.$Values($\bm{l}^\mathrm{meas}_i$\footnote{$\bm{l}^\mathrm{meas}_i = \bm{f}(\bz^\mathrm{meas}_{\bx_i},\bz^\mathrm{meas}_{\theta_i},\bz^\mathrm{meas}_{r_i})$ denotes measured obstacle location.}) \Comment{Add obstacle location to solution}
	\If{$i = 0$} 
		\State $G.\Phi.\mathsf{remove}\left(\phi^{\rm start}(\bx_0)\right)$ \Comment{Remove start factor in \cref{eqn:StartFactor}}
	\ElsIf{$i = N$} 
		\State $G.\Phi.\mathsf{remove}\left(\phi^{\rm goal}(\bx_N)\right)$ \Comment{Remove goal factor in \cref{eqn:EndFactor}}
	\EndIf
	\State $G.\Phi.\mathsf{remove}\left(\phi_i^{\rm lim}(\bu_i),\phi_i^{\rm obs}(\bx_i)\right)$ \Comment{Remove outdated factors}
	\If{$PlanningMode$ = Reactive}  \Comment{Plan given ${\bl}^{\mathrm{meas}}_{i}$}
		\For{$k=i+1:N$} \Comment{Assign cost for future obstacle factors}
			\State $G.\Phi.\mathsf{replace}\left(\phi_k^{\rm obs}\left(\bx_k\right),\bl^{\rm meas}_{i}\right)$ \Comment{Obstacle factor in~\cref{eqn:ObsLike}}
		\EndFor
	\EndIf
	\State $\Theta^*_{i+1} \gets  \mathsf{gtsam}$.LevenbergMarquardt($G,\Theta^*_{i}$).Optimize() \Comment{Solve}
\EndFor
\end{algorithmic}
\end{algorithm}
\vspace*{-22 pt}
\end{minipage}

\section{Experimental Validation} 
\label{sect:Facility}
\begin{figure}[!h]
 	\vspace*{7pt}
	\begin{subfigure}[h]{0.499\linewidth}
		\centering
		\includegraphics[width=1\linewidth]{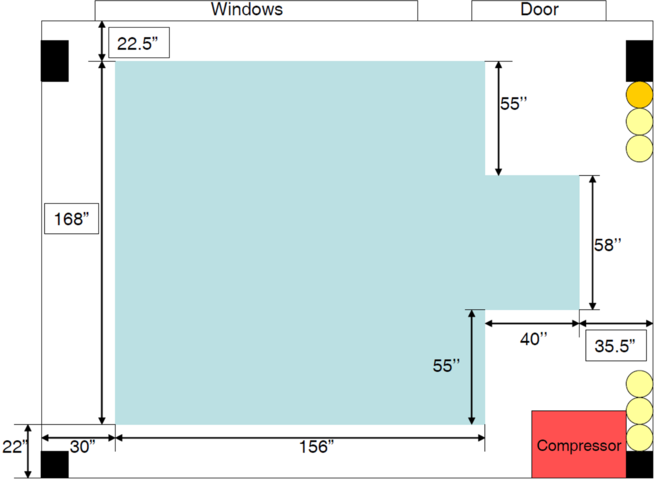}
		\caption{Dimensions of the testing arena.}
		\label{fig:epoxyFloor}
	\end{subfigure}%
	\hfill
	\begin{subfigure}[h]{0.499\linewidth}
		\centering
		\includegraphics[width=0.975\linewidth]{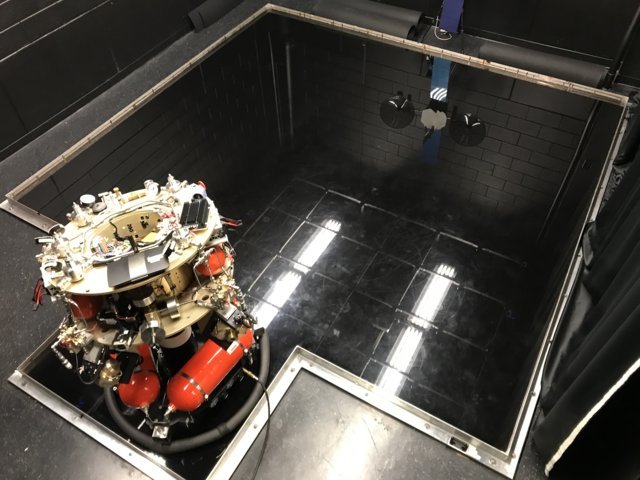}
		\caption{ASTROS platform in arena bay.}
		\label{fig:Overheadview}
	\end{subfigure}
	\caption{The DCSL's robotic air-bearing spacecraft testing facility.}
	\label{fig:ASTROSFacility}
\end{figure}

We have verified SCATE for onboard spacecraft navigation in the Dynamics and Control Systems Laboratory's (DCSL) friction-less robotic spacecraft simulator experimental facility, shown in~\Cref{fig:ASTROSFacility}.
A mounted overhead network of 12 VICON\textsuperscript{TM} cameras monitor a 4m x 4m flat epoxy floor arena shown in~\Cref{fig:Overheadview}, providing localization necessary for inertial pose estimation of all robots within the facility with sub-millimeter and sub-degree
accuracy post calibration~\cite{Tsiotras2014}.
 

\Cref{fig:Overheadview} also shows the 5-DOF air-bearing robotic platform, ASTROS (Autonomous Spacecraft Testing of Robotic Operations in Space)~\cite{Cho2009,Tsiotras2014}, which is equipped with a three-axis inertial measurement unit, a three-axis rate gyro, and
12 pressurized air thrusters, arranged in a 3-3-3-3 configuration, which impart changes in linear and angular momentum necessary for trajectory tracking. 

\subsection{Planar Mechanics and SCATE MAP Trajectory Tracking}

For this analysis and experiment,
we assume that the ASTROS platform is confined to 3-DOF planar air-bearing motion such that the state is
$
	\bm{x}~=~\begin{bmatrix}
		x^{\V} & 
		\dot{x}^{\V} & 
		y^{\V}&
		\dot{y}^{\V}  &
		\psi^{\V/\Sb}&
		\dot{\psi}^{\V/\Sb}
	\end{bmatrix}^{\T}~\in~\R^6,
$
where 
\setlength{\columnsep}{5pt}
\begin{wrapfigure}{r}{0.495\linewidth}
	\vspace*{-8pt}
	\centering
	\includegraphics[width=0.8\linewidth]{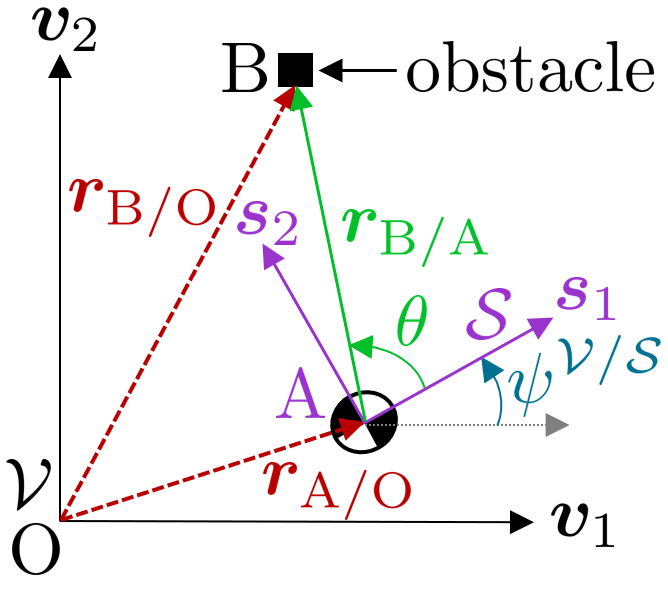}
	\caption{{\color{MidnightBlue}Heading angle}, {\color{MidnightBlue} $\psi^\mathcal{V/S}$}, {\color{ForestGreen} bearing angle}, {\color{ForestGreen} $\theta$}, frames, and center of mass definitions.}
	\label{fig:Kinematics}
	\vspace*{4pt}
\end{wrapfigure}
$x^{\V}$, $y^{\V}$ are the planar components of displacement vector $\bm{r}^{\V}_{\rm {A / O}} = [x^{\V},y^{\V}]^{\T}$, and ${\color{MidnightBlue} \psi^{\V / \Sb}}$ is the {\color{MidnightBlue} heading angle} between $\bm{v}_1$ of the inertial frame $\mathcal{V} = \{{\rm O},\bm{v}_1,\bm{v}_2\}$ and ${\color{DarkOrchid} \bm{s}_1}$ of body-fixed frame ${\color{DarkOrchid}\mathcal{S}} = \{{\color{DarkOrchid} {\rm A},\bm{s}_1,\bm{s}_2}\}$, as shown in~\Cref{fig:Kinematics}. 
The  translational dynamics are 
\begin{equation}
	m\ddot{\bm{r}}^{\V}_{\rm {A / O}} =  ^{\rm A}\hspace{-3pt}\bm{f}^{\V},
	\label{eqn:Netwon}
\end{equation}
where $m$ is the mass, and $^{\rm A}\bm{f}^{\V} = [^{\rm A}f_x^{\V}, ^{\rm A}f_y^{\V}]^\T$ is the applied planar force and the single-DOF rotational attitude dynamics are given by
\begin{equation}
	^{\rm A}I^{\V}_{zz}\ddot{\psi}^{ \V/\mathcal{S}} = ^{\rm A}\hspace{-3pt}\tau^{\V}_z,
	\label{eqn:Euler}
\end{equation}
where $^{\rm A}I^{\V}_{zz}$ is the moment of inertia and $^{\rm A}\tau^{\V}_z$ is the applied torque about the body-fixed z-axis. 
The system can be written in a state-space form (\ref{eqn:SimpleSS}) where
the control input is $\bm{u}= [^{\rm A}f_x^{\V},^{\rm A}f_y^{\V},^{\rm A}\hspace{-2pt}\tau^{\V}_z]^{\T}$. 

We track the SCATE factor graph MAP solution $\Theta^*=\{\hat{\bx}\cup\check{\bx},\hat{\bu} \cup\check{\bu}, \hat{\bl}\}$ to either~\cref{eqn:reactiveSCATE} or~\cref{eqn:PredictiveSCATE}, during online implementation by letting $\bu$ be
\begin{equation}
	\bu= \check{\bu} - K( \bx - \check{\bm{x}}) ,
	\label{eqn:MyopicSCATE_Control}
\end{equation}
and designing a feedback matrix $K \in \R^{3 \times 6}$ such that the matrix $A - BK$ is Hurwitz.  

\subsection{Implementing Online Factor Graph Planning}

The proposed online
factor graph planning is integrated in these experiments 
by using by two computers: one as the control computer, which (re)solves the factor graph given the available information, and an onboard real-time Simulink Speedgoat\textsuperscript{TM} computer, which executes the current plan. 
UDP packets of the set $\{\check{\bx},\check{\bu}\}$ from the SCATE MAP $\Theta^* = \{\hat{\bx} \cup\check{\bx},\hat{\bu} \cup \check{\bu}, \hat{\bl}\}$ are communicated to the onboard computer.
The onboard computer simultaneously solves a linear program~\cite{GLPK} to allocate the onboard thrusters to execute $\bu $ in~\cref{eqn:MyopicSCATE_Control} such that $\bu = \check{\bu} - K(\hat{\bx}_{\mathrm{EKF}}-\check{\bx})$, where $\hat{\bx}_{\mathrm{EKF}}$ is an extended Kalman Filter (EKF) estimate of ASTROS' state~\cite{Filipe2015}, and sends UDP packets of $\hat{\bx}_{\mathrm{EKF}}$, and measurements of the obstacle, $\bz^{\rm meas}$. 
Note that $\hat{\bx}_{\rm EKF}$ is computed at 100 Hz, whereas $\Theta^*$ is computed at 3 Hz, hence $\hat{\bx}_{\rm EKF}$ is used for feedback and is treated as a state measurement during optimization.

\section{Experimental Results} \label{sect:Resuts}

The viability of reactive SCATE factor graph navigation is evaluated experimentally in an environment with an unknown obstacle which is: a) static and b) moving, while
predictive SCATE navigation is tested both on hardware and in software via a to-scale-CAD mesh rendering of the DCSL spacecraft testing facility, ASTROS, and a new 3-DOF air-bearing robot. The objective is for the ASTROS platform to navigate collision-free from rest at one corner of the workspace to a fixed position and attitude in the diagonal corner~\footnote{Results available online at~\href{https://youtu.be/C03TKAtqm8I}{https://youtu.be/C03TKAtqm8I}.}.

\subsection{Reactive SCATE Navigation With A Static Obstacle}

\begin{figure}[!h]
	\begin{subfigure}[h]{0.32\linewidth}
		\centering
		\includegraphics[width=1\linewidth]{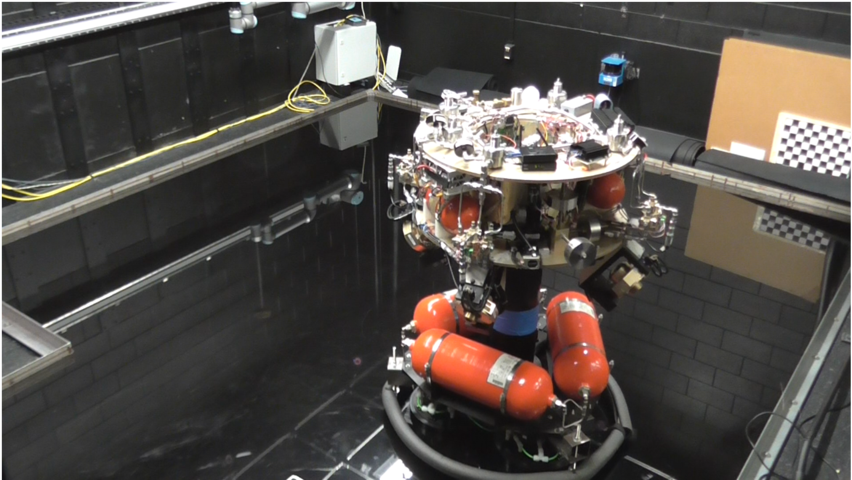}
		\caption{Time: $t=0$s.}
		\label{fig:Static1}
	\end{subfigure}%
	\hfill
	\begin{subfigure}[h]{0.32\linewidth}
		\centering
		\includegraphics[width=1\linewidth]{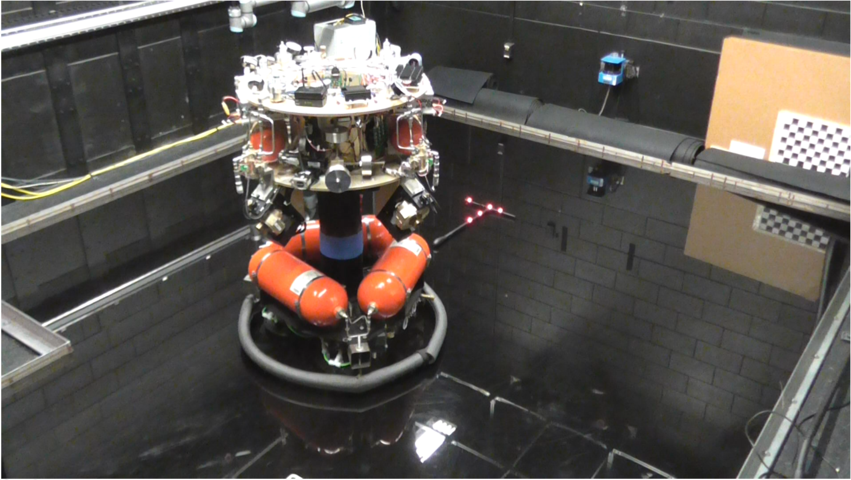}
		\caption{Time: $t=20$s.}
		\label{fig:Static2}
	\end{subfigure}
	\hfill
	\hspace*{-7pt}
	\begin{subfigure}[h]{0.32\linewidth}
		\centering
		\includegraphics[width=1\linewidth]{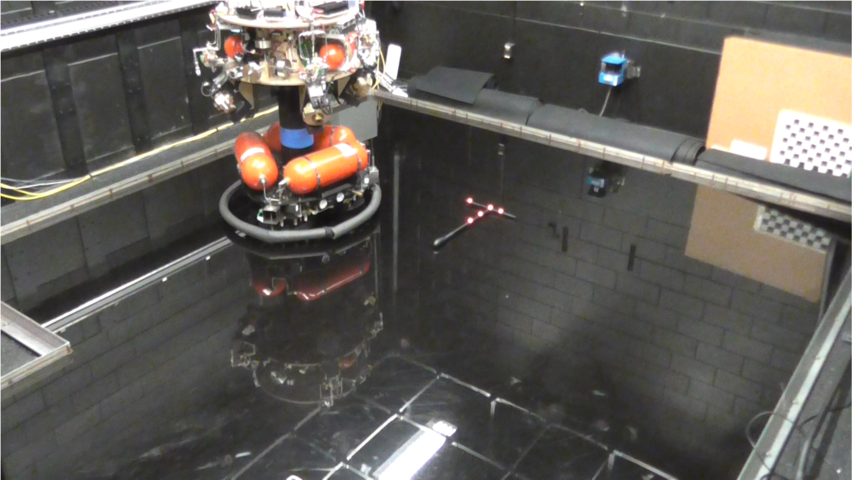}
		\caption{Time: $t=60$s.}
		\label{fig:Static3}
	\end{subfigure}
	\caption{Reactive SCATE navigation in a static obstacle environment.}
	\label{fig:SCATEstaticHardware}
\end{figure}

\begin{figure}[!ht]
 	\vspace*{-5pt}
	\centering
	\includegraphics[width=0.975\linewidth]{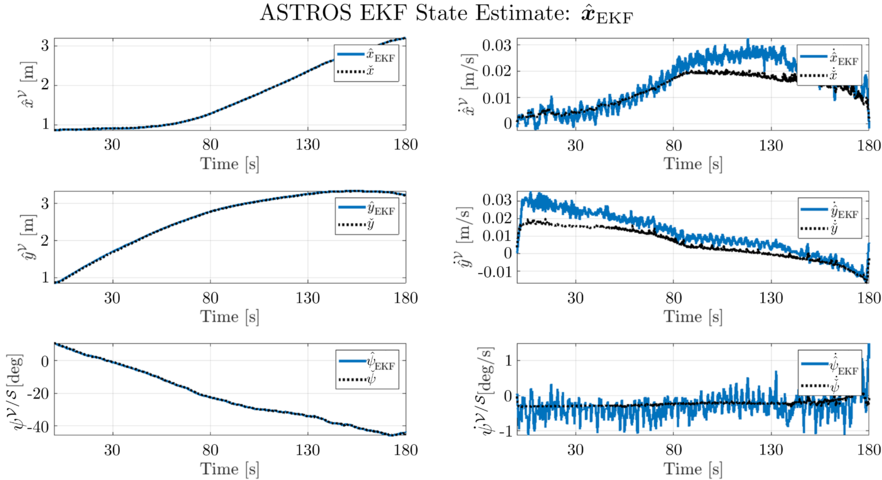}
	\caption{State estimate, $\hat{\bx}_{\rm EKF}$, and MAP reference, $\check{\bx}$.}
	\label{fig:staticHWstate}
\end{figure}

Reactive SCATE navigation is tested with a static obstacle (i.e., the VICON\textsuperscript{TM} wand) environment, as shown in~\Cref{fig:SCATEstaticHardware}. 
The system EKF state estimate, $\hat{\bx}_{\rm EKF}$, is given in~\Cref{fig:staticHWstate}, where along with ~\Cref{fig:SCATEstaticHardware}, we see that the ASTROS platform navigates collision-free along the MAP reference trajectory, $\check{\bx}$, reaching the terminal goal state. 

\subsection{Reactive SCATE Navigation With A Moving Obstacle}

\begin{figure}[!h]
	\begin{subfigure}[h]{0.32\linewidth}
		\vspace*{8pt}
		\centering
		\includegraphics[width=1\linewidth]{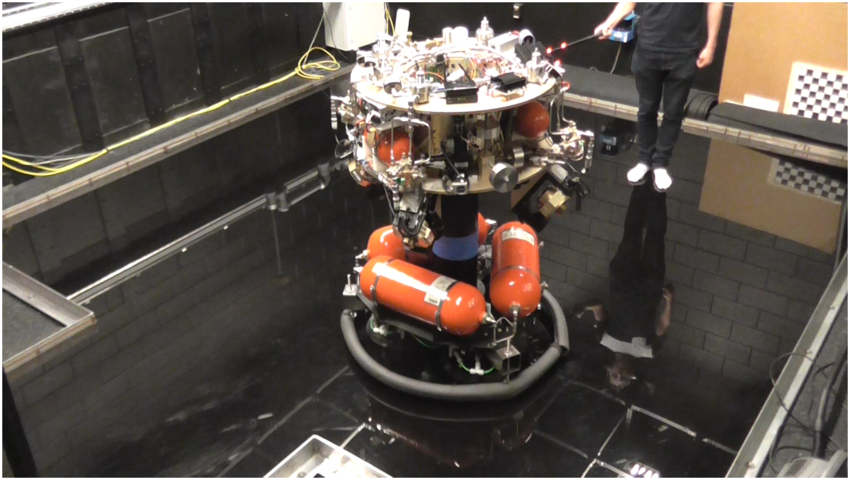}
		\caption{Time: $t=0$s.}
		\label{fig:Dync1}
		\vspace{8pt}
	\end{subfigure}%
	\hfill
	\begin{subfigure}[h]{0.32\linewidth}
		\centering
		\includegraphics[width=1\linewidth]{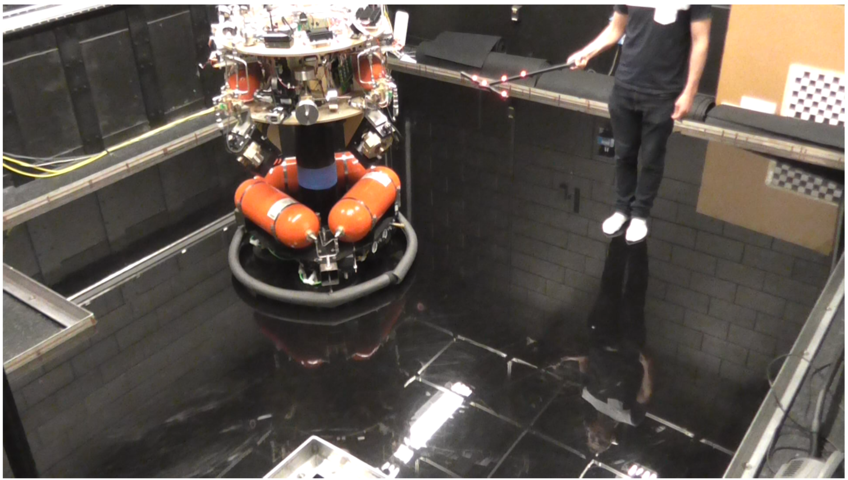}
		\caption{Time: $t=40$s.}
		\label{fig:Dync4}
	\end{subfigure}%
	\hfill
	\begin{subfigure}[h]{0.32\linewidth}
		\centering
		\includegraphics[width=1\linewidth]{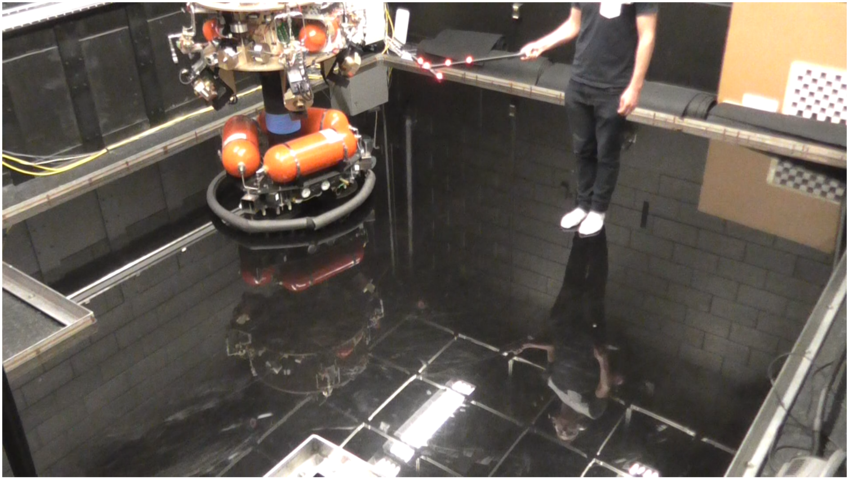}
		\caption{Time: $t=80$s.}
		\label{fig:Dync6}
	\end{subfigure}
	\caption{Reactive SCATE navigation with a moving obstacle.}
	\label{fig:DyncEvl}
\end{figure}

We now move the obstacle around the workspace, forcing the ASTROS  platform to navigate to the perimeter of the arena in order to avoid an obstacle collision while navigating to the terminal reference waypoint, as shown in~\Cref{fig:DyncEvl}. 
\begin{figure}[!ht]
	\centering
	\includegraphics[width=1\linewidth]{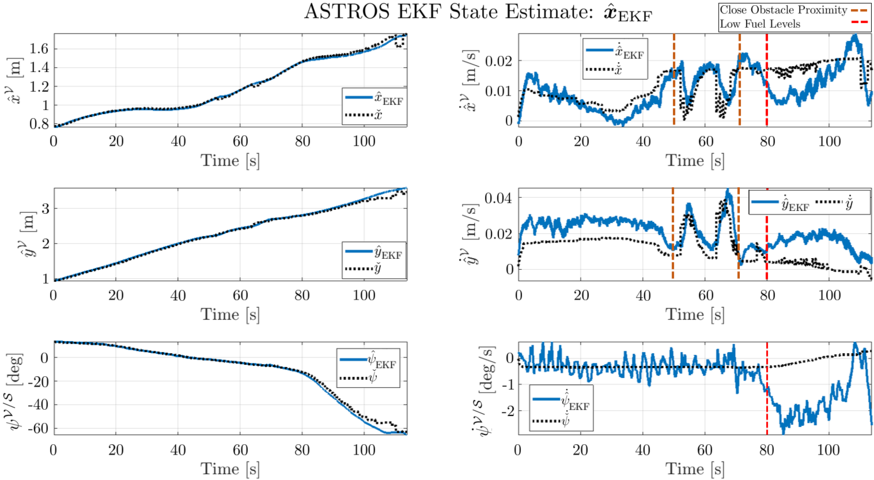}
	\caption{State estimate, $\hat{\bx}_{\rm EKF}$, and MAP reference, $\check{\bx}$, during reactive SCATE navigation around a moving obstacle.}
	\label{fig:dycHWstate}
	\vspace*{-3pt}
\end{figure}

Note that the large instantaneous changes in the MAP reference planar velocities during the 45-70 seconds is the result of close obstacle proximity, as shown in~\Cref{fig:dycHWstate}. 
After 80 seconds, 
Figure~\ref{fig:dycHWstate} shows that the deviation from the MAP reference trajectory increases, which is a result of the exhausting the fuel necessary for proper thruster allocation.  

\subsection{Predictive SCATE Navigation With A Moving Obstacle}

\subsubsection{Hardware Validation} 

\begin{figure}[!h]
	\begin{subfigure}[h]{0.32\linewidth}
		\centering
		\includegraphics[width=1\linewidth]{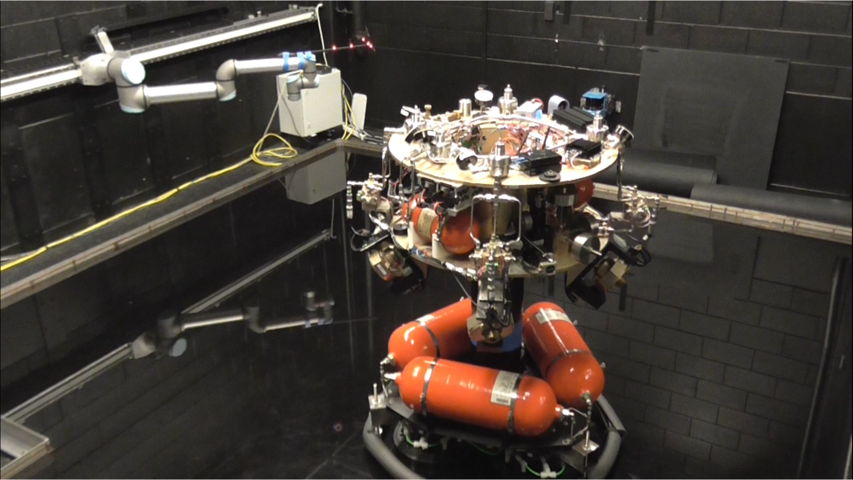}
		\caption{Time: $t=0$s.}
		\label{fig:Pred1}
	\end{subfigure}%
	\hfill
	\begin{subfigure}[h]{0.32\linewidth}
		\centering
		\includegraphics[width=1\linewidth]{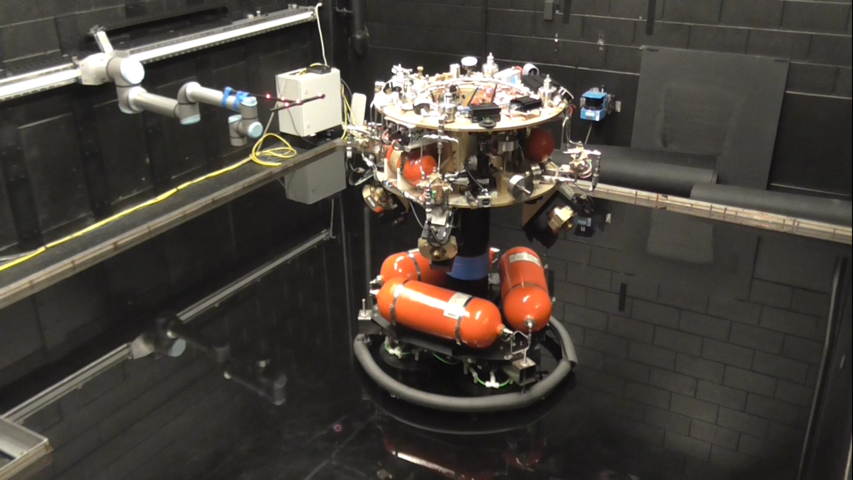}
		\caption{Time: $t=40$s.}
		\label{fig:Pred2}
	\end{subfigure}
	\hfill
	\hspace*{-7pt}
	\begin{subfigure}[h]{0.32\linewidth}
		\centering
		\includegraphics[width=1\linewidth]{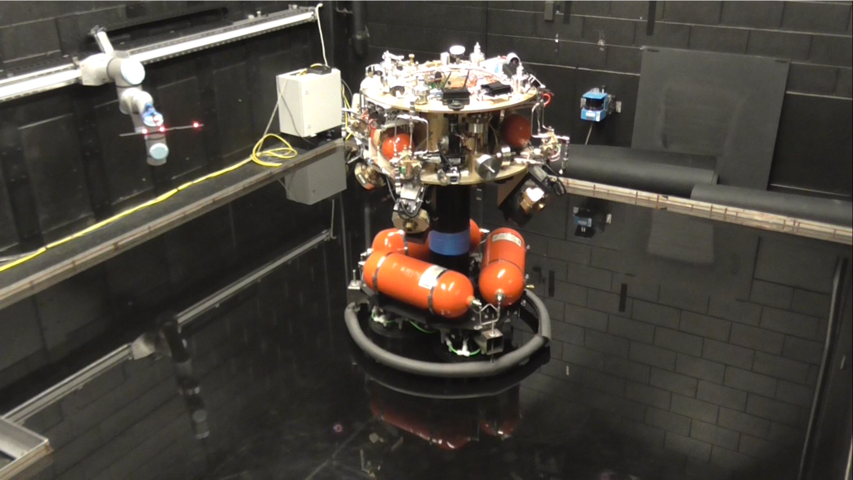}
		\caption{Time: $t=60$s.}
		\label{fig:Pred3}
	\end{subfigure}
	\caption{Predictive SCATE navigation with a moving obstacle.}
	\label{fig:SCATEPredHardware}
\end{figure}

\begin{figure}[!ht]
    \vspace*{-10pt}
	\centering
	\includegraphics[width=1\linewidth]{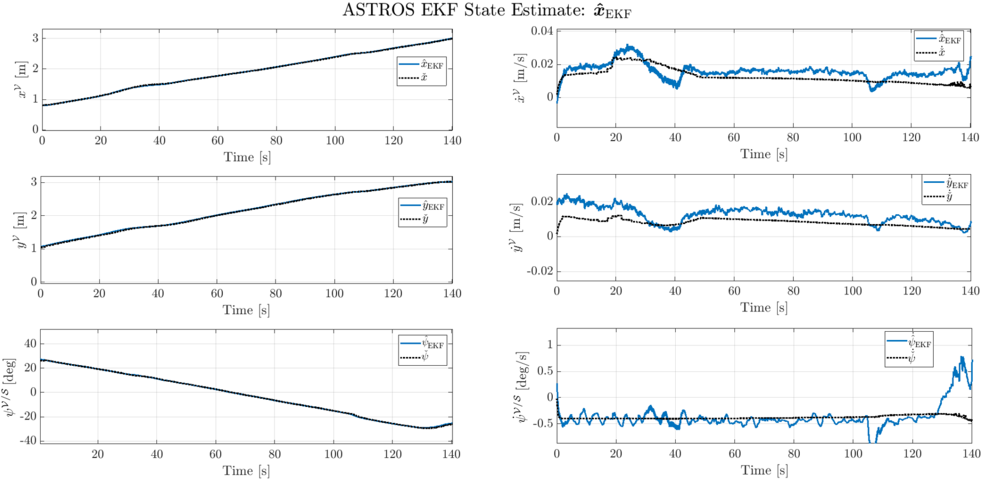}
	\caption{State estimate, $\hat{\bx}_{\rm EKF}$, and MAP reference, $\check{\bx}$, during predictive SCATE navigation around a moving obstacle.}
	\label{fig:PredHWstate}
\end{figure}

For predictive SCATE implementation, given in~\cref{eqn:PredictiveSCATE}, we attached the obstacle to a 7-DOF UR10e\textsuperscript{TM} manipulator such that the obstacle follows a predefined trajectory, i.e., $\tilde{\bl}$. The ASTROS platform navigates collision-free around the obstacle, as shown in~\Cref{fig:SCATEPredHardware},
and reaches the terminal goal state with small deviations from the MAP reference trajectory, as shown in~\Cref{fig:PredHWstate}.

\subsubsection{Numerical Simulation}
\begin{figure}[!ht]
	\centering
	\includegraphics[width=0.975\linewidth]{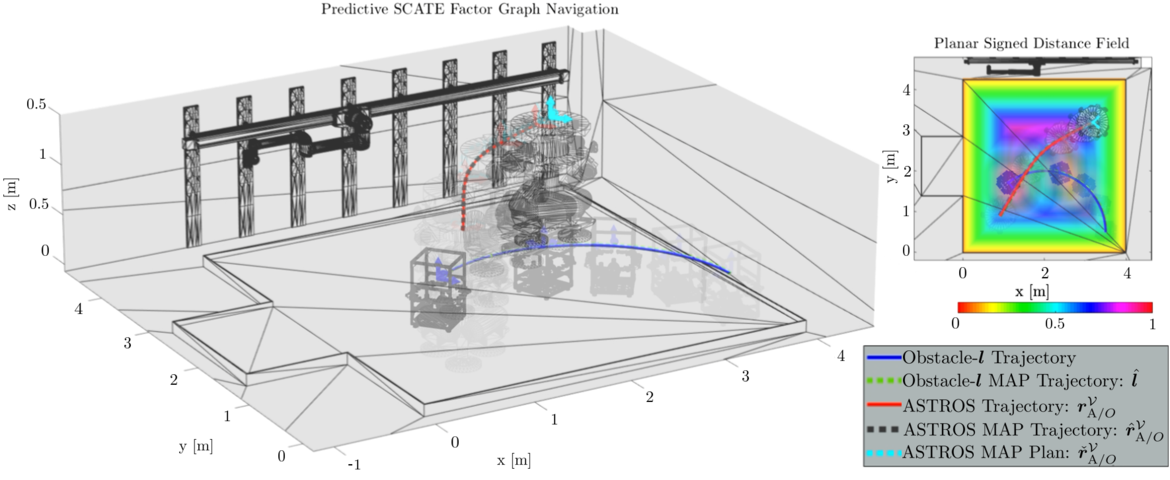}
	\caption{Predictive SCATE navigation. Faded figures show the past.}
	\label{fig:PredSCATEevolution}
\end{figure}

\begin{figure}[!ht]
	\begin{subfigure}[h]{0.49\textwidth}
	\centering
	\includegraphics[width=0.975\linewidth]{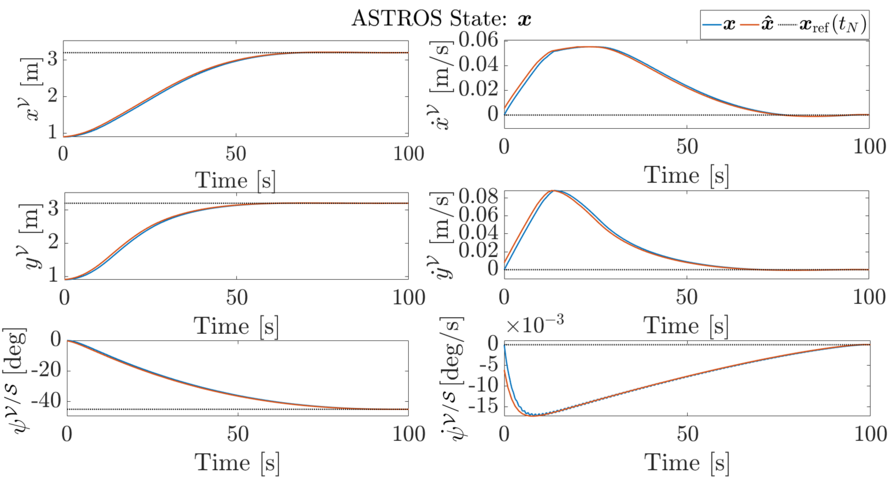}
	\caption{Platform state, $\bx$, MAP estimate, $\hat{\bx}$, and reference, $\bx_{\rm ref}(t_N)$. }
	\label{fig:PredSCATE_StateErr}
	\vspace{8pt}
	\end{subfigure}%
	\hfill
	\begin{subfigure}[h]{0.49\textwidth}
	\centering
	\includegraphics[width=0.98\linewidth]{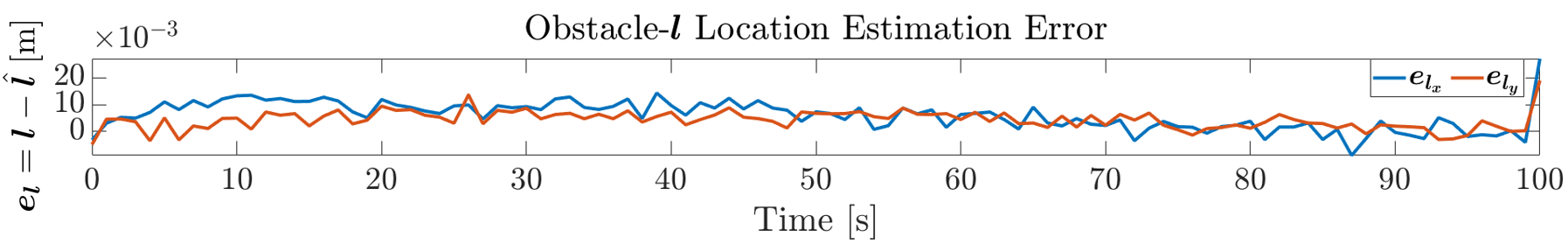}
	\caption{MAP obstacle trajectory estimation error, $\bm{e}_{\bl} = \bl - \hat{\bl}$.}
	\label{fig:PredSCATE_ObsErr}
	\end{subfigure}%
	\caption{Predictive SCATE simulation trajectories and estimations.}
	\label{fig:PredSCATEEsts}
\end{figure}

Predictive SCATE factor graph navigation is also simulated with the same navigation constraints as the reactive SCATE case, with another robot, as shown in~\Cref{fig:PredSCATEevolution}. 
The ASTROS platform successfully navigates without collisions around the obstacles to reach the terminal state, as shown in~\Cref{fig:PredSCATEevolution,fig:PredSCATE_StateErr}. 
Finally, \Cref{fig:PredSCATEEsts} shows that the MAP estimates  $\hat{\bx}$ and $\hat{\bl}$, are in close correspondence with $\bx,\bl$, respectively.

\section{Conclusion} \label{sect:Conclusion}

We show the utility of using factor graphs for simultaneous control and trajectory estimation (SCATE) for collision-free navigation of autonomous robotic spacecraft systems in environments with moving objects with and without an obstacle motion model. 
We have numerically and experimentally validated SCATE factor graph navigation, confirming the algorithm's capacity to: a) incorporate realistic vehicle dynamics, b) generate online collision-free reference trajectories and estimates of vehicle state, control, and obstacle trajectories in a moving obstacle environment, and c) navigate spacecraft systems autonomously in support of future OSS missions.

Although the utility of this algorithm has only been demonstrated for linear planar mechanics, SCATE factor graph navigation is readily extendable to 6-DOF nonlinear single- and multi-body dynamics, as well as, 3D motion planning, where 3D SDFs are necessary. 
This investigation is part of ongoing and future work.

\bibliographystyle{IEEEtran}
\bibliography{IEEEabrv,FactorLib}

\end{document}